\documentclass[sn-mathphys]{sn-jnl}

\theoremstyle{thmstyleone}%
\theoremstyle{thmstyletwo}%

\theoremstyle{thmstylethree}%

\twocolumn

\begin{document}

\title[Object Detector Differences when using Synthetic and Real Training Data]{Object Detector Differences when using Synthetic and Real Training Data}


\author*[1]{\fnm{Martin Georg} \sur{Ljungqvist}}\email{martin.ljungqvist@axis.com}  

\author[1]{\fnm{Otto} \sur{Nordander}}

\author[1]{\fnm{Markus} \sur{Skans}}\email{markus.skans@axis.com}

\author[2]{\fnm{Arvid} \sur{Mildner}}

\author[2]{\fnm{Tony} \sur{Liu}}

\author*[2]{\fnm{Pierre} \sur{Nugues}}\email{pierre.nugues@cs.lth.se} 

\affil*[1]{\orgname{Axis Communications AB}, \orgaddress{\city{Lund}, \country{Sweden}}}

\affil[2]{\orgdiv{Department of Computer Science}, \orgname{Lund University}, \orgaddress{\city{Lund}, \country{Sweden}}}

\abstract{
    To train well-performing generalizing neural networks, sufficiently large and diverse datasets are needed. Collecting data while adhering to privacy legislation becomes increasingly difficult and annotating these large datasets is both a resource-heavy and time-consuming task. An approach to overcome these difficulties is to use synthetic data since it is inherently scalable and can be automatically annotated. However, how training on synthetic data affects the layers of a neural network is still unclear. In this paper, we train the YOLOv3 object detector on real and synthetic images from city environments. We perform a similarity analysis using Centered Kernel Alignment (CKA) to explore the effects of training on synthetic data on a layer-wise basis. The analysis captures the architecture of the detector while showing both different and similar patterns between different models. With this similarity analysis we want to give insights on how training synthetic data affects each layer and to give a better understanding of the inner workings of complex neural networks. The results show that the largest similarity between a detector trained on real data and a detector trained on synthetic data was in the early layers, and the largest difference was in the head part. The results also show that no major difference in performance or similarity could be seen between frozen and unfrozen backbone.
}

\keywords{Object Detection, Layer Similarity, Centered Kernel Alignment}



\maketitle

\section{Introduction}
\label{sec:introduction}

Using convolutional neural networks (CNNs) is a popular approach to solve the object detection problem in computer vision. A lot of effort has been put into developing accurate and fast object detectors leveraging the structure of convolutional layers \cite{ssd,retinanet,yolov3,efficientdet2020}. This has led to a drastic increase in performance of object detectors during the past few years. However, these models generally require massive amounts of labeled training data to achieve good performance and generalization \citep{how-much-data}. Building these datasets can be both time consuming and resource heavy.

First, the raw data need to be collected, often involving complex data acquisition setups and gathering schemes. Adhering to privacy, data protection regulations and ensuring the diversity and quantity of the data becomes an increasingly difficult challenge.

Second, the data need to be annotated. Since datasets for deep learning often include several thousand images, the annotation process becomes a very mundane, time-consuming, and error-prone task.

One way of avoiding these issues is using synthetic data for training. Generated synthetic datasets are inherently scalable and labelling of the data can be done automatically. These datasets can for example be generated using a data generation tool such as \textit{Carla} \citep{carla}, or sampling videos from open-world video games like \textit{Grand Theft Auto V (GTA V)} \citep{richter2017benchmark,gtav-matrix}.

A general problem with deep neural networks is that their complexity makes it difficult to understand exactly why a certain prediction has been made. This has led to neural networks often being considered as black boxes \citep{blackbox2,blackbox}, where one only looks at the input and the output, while relying on trial and error when creating a well-working system. CNNs are less regarded as black boxes since they are suitable for visualisation, but that renders a vast amount of information to overview and may not tell everything about the networks inner workings. There have been many studies on understanding and visualizing the inner workings of deep convolutional neural networks \citep{zeiler2014,blackbox2,blackbox,svcca,pwcca,cka,alllayersequal,hermann2020,nguyen2021cka,interpreting2021}. For example, \cite{blackbox2} showed that classification error decreased monotonically for each layer in a trained ResNet model, indicating that each layer contributed to the end result. \cite{zeiler2014} visualized both filters and activations in a small CNN.

In this work, we investigate how object detection models are affected when trained on synthetic data versus real data by exposing the inner workings of the network. One key element will be the comparison between the outputs from individual hidden layers in the models using the recently proposed idea of similarity measurement \citep{cka}. Our work builds upon \cite{liumildnerthesis} and is an extended version of \cite{ljungqvist2022} with further results and more detailed analysis.

Our aim is to investigate how synthetic data affects the performance of object detection models as well as how hidden layers in the CNNs are affected by different types of training data. More specifically:
\begin{enumerate}[1.]
    \item How does a model trained on synthetic data differ from one trained on real data and what network layers are affected?
    \item Does freezing the backbone affect this? 
\end{enumerate}
To the best of our knowledge, no such analysis has been made on a CNN object detector using real and synthetic data.

Our main contributions are:
\begin{itemize}
    \item We show what parts of the network are most similar for a detector trained on real image data compared to when it is trained on synthetic data.
    \item We also determine the consequences of freezing the backbone or not when further training a detector on synthetic data.
\end{itemize}

\section{Related Work}
\label{sec:related_work}
\subsection{Object Detection}
Two-stage object detection models use a region-based detection process. The first stage proposes a sparse set of candidate object locations, and the second stage classifies the candidate as either a background or foreground object. Network architectures such as Faster RCNN \citep{Faster-RCNN}, and FPN \citep{FPN} use this two-stage process.

One-stage detectors are suitable for use in real-time object detection in video. These methods sample densely on the set of object locations, scales, and aspect ratios. Proposed methods are for example YOLO \citep{yolo}, RetinaNet \citep{retinanet}, SSD \citep{ssd} and EfficientDet \citep{efficientdet2020}. These networks are significantly faster while having comparable performance to the conventional two-stage methods. Because of its speed, comparable accuracy, and relatively light-weightness, YOLOv3 \citep{yolov3} was chosen for our experiments.

\subsection{Synthetic Data}
There are several synthetic datasets of city environments available and several experiments of training on synthetic data have been conducted.

VKITTI \citep{vkitti,vkitti2} is a synthetic version of the KITTI dataset \citep{kitti}, but it does not contain persons. Synthia \citep{synthia} is another synthetic dataset of images from urban scenes, where the results showed increased performance when training on a mixture of real and synthesized images. The video game GTA V has been used to generate synthetic datasets \citep{richter2017benchmark,gtav-matrix}.

The experiments conducted in \cite{gtav-matrix} showed that training a Faster R-CNN on a GTAV synthetic dataset of at least 50,000 images increased the performance compared to training on the smaller real dataset Cityscapes \citep{cityscapes} when evaluated on the real KITTI dataset \citep{kitti}. However, these experiments only used cars as labels, disregarding other labels such as persons and bicycles.

The Synscapes dataset is a synthetic version of Cityscapes \citep{synscapes}. The authors claim that training on only Synscapes yields decent results, but lowers performance compared to training on real data when evaluated on Cityscapes. However, their experiments showed that models trained on Synscapes outperformed both models trained only on GTAV \citep{richter2017benchmark} and Synthia \citep{synthia}.

Furthermore, \cite{synscapes} claimed that training on a mixture of synthetic and real data can further improve performance, outperforming models trained only on real data. Results from \cite{how-much-data} showed that training on synthetic data and fine-tuning on real data yielded better performance than training on a mixed real-synthetic dataset. The authors also concluded that photo-realism in the synthetic data was not necessarily as important as other factors in the training such as diversity.

Non-artistically generated images have been produced by \textit{domain randomization} \citep{tremblay2018nvidia}, where parameters such as lighting, pose, and textures were randomized. The authors showed that with additional fine-tuning on real data, their model outperformed models trained only on real data for object detection of cars on the KITTI dataset. Furthermore, they argued that letting the backbone be trainable during training on synthetic data yielded better performance compared to freezing the backbone weights.

Synthetic data have been used for pedestrian detection and pose estimation \citep{hattori2018}. The authors showed that training on synthetic images only yielded a model that outperformed a model trained on real data only. However, the models were scene-specific and location-specific where they used a priori knowledge about the camera parameters and the scene geometry.

The authors of \cite{hinterstoisser2018buzz} superimposed 3D rendered models of toys with different lighting and poses onto real backgrounds. As opposed to \cite{tremblay2018nvidia}, the authors argued that freezing backbone weights (when they are initialized from a pre-trained backbone) during training on the synthetic data yielded better performance compared to letting the backbone be trainable. The authors of \cite{tremblay2018nvidia} argued that a possible explanation could be that the dataset that they used was large and diverse enough to further improve the backbone.

In the work by \cite{astermark2018} synthetic data of faces was generated using Generative Adversarial Networks (GAN). The author showed that while training on large real training sets yielded best performance, synthetic data can improve performance if small amounts of real data are available.

Furthermore, \cite{harrysson2019} trained a YOLOv3 detector on synthetic license plates superimposed on real background images. The results showed that mixing real and synthetic data gave better performance and only using synthetic data for training almost matched the performance of training on only real data.

\subsection{Similarity of Neural Networks}
One way of obtaining more insight on how a CNN network behaves is looking at the outputs layer-wise. By comparing layer outputs from two different models, one can determine the similarity between the layers. One method of measuring the similarity of layer outputs is the \textit{singular value canonical correlation analysis} (SVCCA) \citep{svcca}. SVCCA uses \textit{singular value decomposition} (SVD) \citep{svd} for dimensionality reduction and then \textit{canonical correlation analysis} (CCA) \citep{cca} which was previously used to learn semantic representations for web images. A further improvement of SVCCA is the \textit{projection weighted CCA} (PWCCA) \citep{pwcca}, which uses projection weighting to calculate the similarity measure as a weighted mean instead of a naive mean as in SVCCA. 

Both metrics are invariant to invertible linear transformations which according to \cite{cka} leads to some major issues. \cite{cka} instead proposed a metric called \textit{centered kernel alignment} (CKA) which, according to the authors, better captures similarity representations between network layers. 

Later work \citep{svcca,pwcca} have shown that the Euclidean distance is not an ideal measurement of similarity between hidden layer outputs, but it can still give some useful insights.

While there exist several papers that attempt to answer how initialization, model complexity, or dataset size affect the similarity between models \citep{svcca,pwcca,cka}, no attempts have been made to compare the difference between models trained on synthetic and real data. As CKA gives a layer-wise similarity of hidden layers within the network, it can give insights of how such networks differ from each other on a layer-basis. These insights could be leveraged for example during training to target specific layers inside networks to improve performance.

\section{Materials and Methods}
\label{sec:materials_and_methods}
\subsection{Datasets}
\subsubsection{Berkeley Deep Drive}

\begin{figure}[h]
  \centering
  \includegraphics[width=\linewidth]{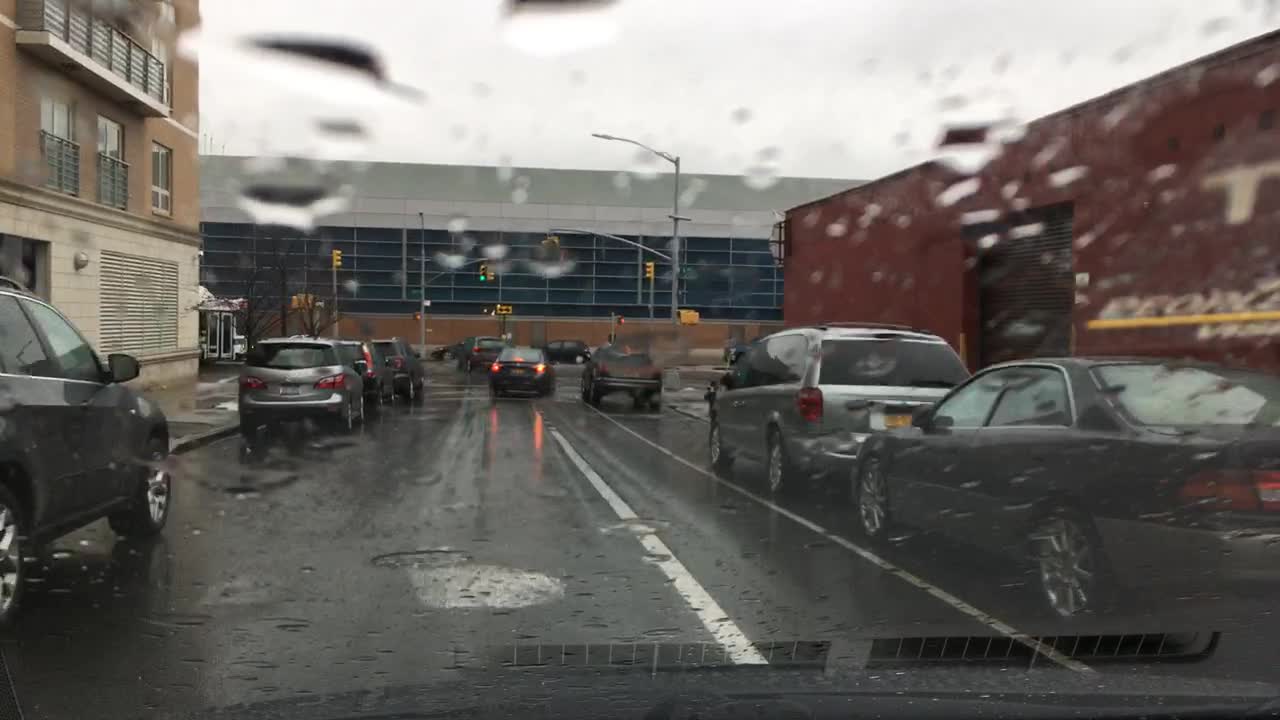}
  \caption{Example image from the BDD dataset \citep{BDD} with vehicles in rain.}
  \label{fig:bdd_dataset}
\end{figure}

The Berkeley Deep Drive (BDD) dataset \citep{BDD} consists of 100,000 driving images collected from 50,000 rides, with $1280 \times 720$ resolution. The images were collected from diverse scenes such as cities, residential areas, and highways, recorded during different hours of the day and in different weather conditions. An example image can be seen in Figure \ref{fig:bdd_dataset}. The images have annotations with bounding boxes and class label.

20,000 out of the 100,000 images are reserved for the test set where the labels are unavailable. Therefore we use only the remaining 80,000 images for our experiments randomly divided into 60/20/20\% for training, validation, and testing\footnotemark. The object mean bounding box size for the objects used is about $120 \times 105$ pixels.

\subsubsection{Grand Theft Auto V}
The Playing for Benchmarks dataset, here denoted GTAV, consists of images of size $1920 \times 1080$ sampled from video sequences from the video game \textit{Grand Theft Auto V} \citep{richter2017benchmark}. Each rendered image has information about the objects' labels and positions.

\begin{figure}[h]
  \centering
  \includegraphics[width=\linewidth]{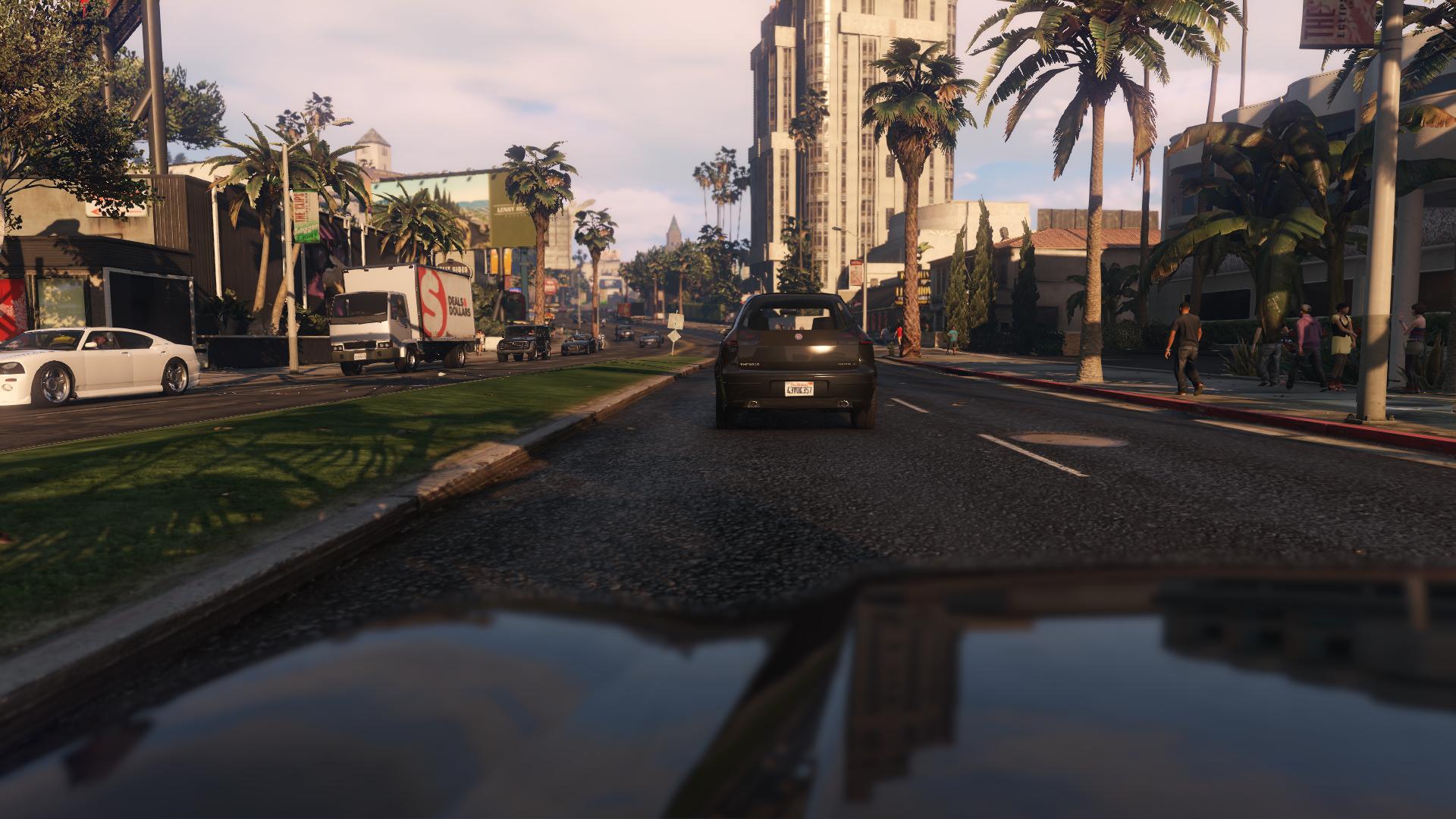}
  \caption{Example image from the GTAV dataset \citep{richter2017benchmark} with vehicles and persons.}
  \label{fig:gtav_dataset}
\end{figure}

The training set consists of about 134,000 street view images which were collected on different time of day, in different weather conditions in a fictional city. An example image can be seen in Figure \ref{fig:gtav_dataset}. Those images were here divided into 60/20/20\% for training, validation, and testing, for the experiments\footnotemark[\value{footnote}]. The object mean bounding box size for the objects used is about $140 \times 120$ pixels.

\footnotetext{https://github.com/ljungqvistmartin/datasplits}

The GTAV dataset consists of labels of objects that can be very far away or persons inside vehicles which makes them very hard or sometimes impossible to spot. Therefore, small bounding boxes with an area smaller than 100 pixels were filtered out. This area was chosen by empirical visual inspection of the ground-truth bounding boxes.

Furthermore, in the GTAV dataset, the hood of the driving car is labeled while it is not labeled in the BDD dataset. Therefore, the hood annotations were not used here.

\subsection{Intersection of Class Labels}
The GTAV \citep{richter2017benchmark} and BDD \citep{BDD} datasets use different class labels. GTAV has 32 classes while BDD has 10, and label names in the datasets differ.

Therefore, a common subset of five classes was selected. This label space is called the \textit{common} labels: car, person, cycle, truck, bus. Most of these objects can be seen in Figures \ref{fig:bdd_dataset} and \ref{fig:gtav_dataset}.

Because of this, the more detailed labels \textit{van} and \textit{trailer} were mapped into the less descriptive common labels \textit{car} and \textit{truck} respectively. Also, motorcycle and bicycle were mapped to the same class. This renders a common interface between the model and the datasets which is independent of which dataset the model is trained on.

The mapping from BDD and GTAV labels to the common labels are shown in Table \ref{tab:bddgtav2common}.

\begin{table*}[ht!]
 \caption{The label map between \textit{BDD}, \textit{GTAV} labels and the  \textit{common} labels.}
    \begin{center}
    \begin{tabular}{ |p{3cm}|p{3cm}|p{3cm}| }
    \hline
\textbf{Common}&\textbf{BDD}&\textbf{GTAV}\\
\hline
  person & person, rider & person\\
  cycle & bike, motor & bicycle, motorcycle \\
  car & car & car, van\\
  bus & bus & bus\\
  truck & truck & truck, trailer\\
\hline
\end{tabular}
\end{center}
\label{tab:bddgtav2common}
\end{table*}

\subsection{YOLOv3}
\label{sec:yolov3}
YOLOv3, \textit{You Only Look Once version 3}, \citep{yolov3} is a one-stage object detector. Compared to similar performing object detection methods, YOLOv3 claims to be faster at inference due to its one-stage detection process. The high inference speed is especially attractive in a real-time detection application.

\begin{figure*}[t]
    \centering
    \includegraphics[width=\linewidth]{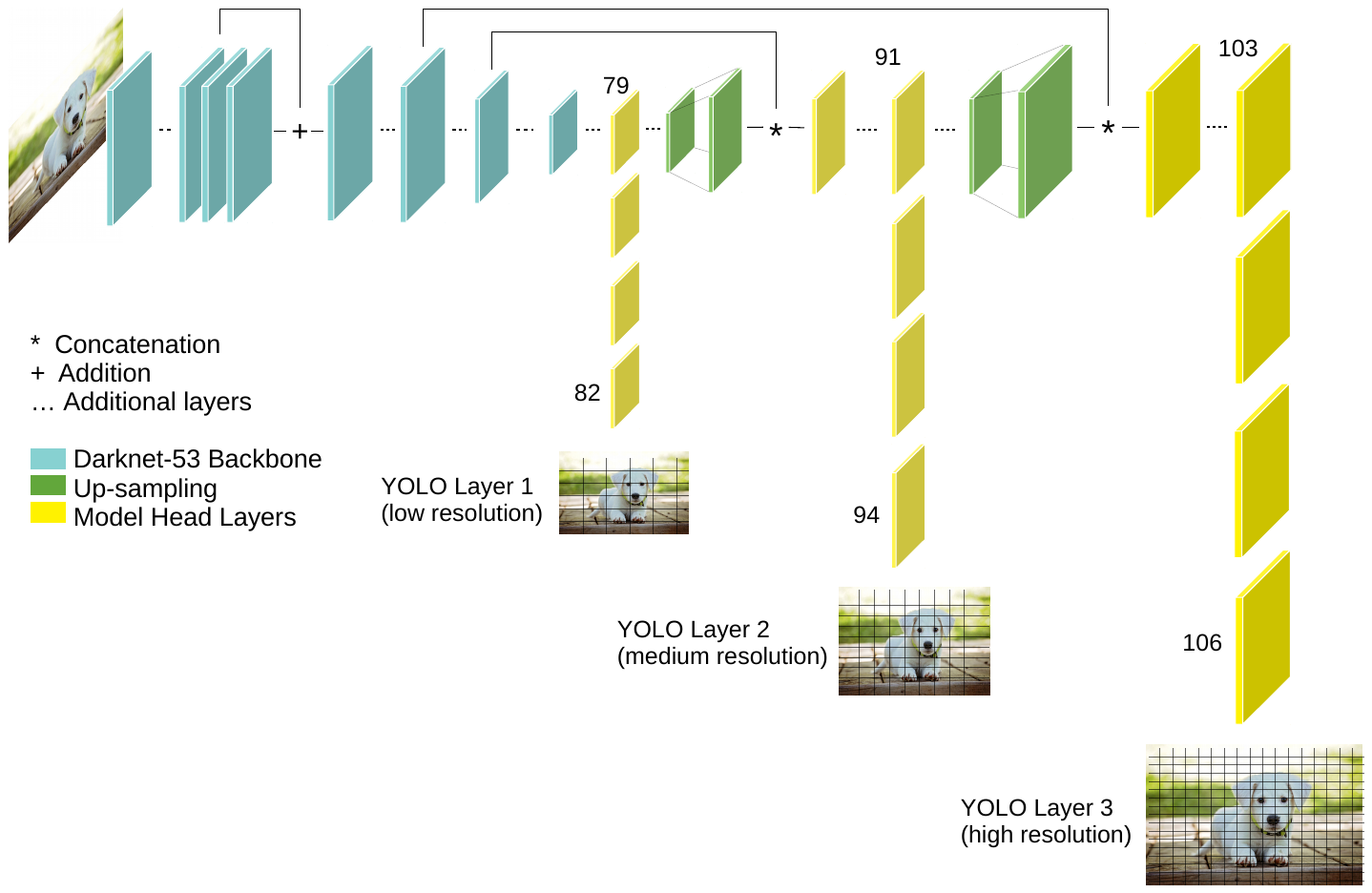}
    \caption{YOLOv3 architecture overview showing the schematic structure of the backbone as well as the head part with the three YOLO detection layers with different resolution. The shortcut in residual blocks and the shortcuts between the backbone and the head are illustrated by lines. Figure from \cite{liumildnerthesis}.}
    \label{fig:yolo_architecture}
\end{figure*}

Figure \ref{fig:yolo_architecture} shows an overview of the YOLOv3 network architecture.

The YOLOv3 architecture builds on extracting features from an image using Darknet-53, a backbone built of 23 residual blocks including 52 convolutional layers, which down-samples along the network depth using the stride length instead of max pooling.

The backbone is divided into residual blocks that are leveraging shortcut connections similarly to ResNet backbones \citep{ResNet}. The benefit of such skip connections is that they deal with vanishing gradients and at the same time encourage feature reuse, which makes the model more parameter-efficient.

The YOLOv3 network contains 107 layers in total (numbered 0 to 106), of which 75 are convolutional layers, 23 residual (shortcut) layers (all in the backbone), 4 route layers where shortcuts end up (all in the head). Downscaling is done by a factor of two at layers 1, 5, 12, 37, and 62 in the backbone. Of the convolutional layers, 38 have a kernel of $3 \times 3$ and 37 have a kernel of $1 \times 1$.

The YOLOv3 network predicts bounding boxes at three resolution levels. These final prediction layers are referred to as detection layers; layers 82, 94, and 106. Each detection layer consists of a grid, where each cell contains the prediction of a bounding box, its objectness score, and a classification score for each class. All three detection layers are immediately preceded by seven convolutional layers.

After the low-resolution detection layer (layer 82), responsible for detecting high-level objects, the output is up-sampled (85) and concatenated (83 and 86) with intermediate output from Darknet-53 (61), which corresponds to the same up-sampled resolution. This concatenated tensor is passed through seven convolutional layers (87-93) and finally through the second detection layer (94). The same procedure is then repeated for the layers preceding the third and last detection layer (106).

The detection layers are grids, where the cells are responsible for predicting the bounding boxes as well as containing the predicted object and class probability. In inference, bounding boxes are non-maximum suppressed according to their objectiveness score, filtering out instances which the network believes have low probability of containing objects. The remaining bounding box predictions are then used in the actual prediction of the model.

\subsection{The Models}
\label{sec:models}
CNNs are often divided into two parts: a backbone responsible for feature extraction and a detection head or classifier. Since training a backbone can be time consuming, training of CNNs often uses pre-trained backbone weights at initialization to reduce the computations needed. It is also advantageous for generalization.

The feature extraction layers could be considered general enough and that it is beneficial to freeze the layers as a kind of regularization \citep{hinterstoisser2018buzz}. On the other hand, the feature extraction layer weights may still have room for actual improvement and further training could increase the overall performance. Therefore, we analyze three differently trained models.

All trained models use the same hyperparameters: a learning rate of $10^{-4}$, a batch size of 8, the Adam optimizer, 100 epochs with patience 10 (early stopping).

The random seed was set to the same value for all training sessions for them to have the same initialization, prerequisites, and for the results to be reproducible. All trainings and analyses were performed twice: initialised with random seed 0, and initialised with random seed 1. Even though CKA can show similarities invariant of the network initialization, the results of \cite{cka} still show small differences. Same initialization seed was used here to make sure that most of the CKA similarity is due to the different datasets. Using two different seeds makes it possible to compare the results also for different initializations.

The images were scaled to $416 \times 416$ pixels for training, test, and analysis. However, the CKA comparison analysis was performed feeding images rescaled to $32 \times 32$ pixels to the networks to make the large matrices of concatenated activations fit in the working memory. Even though the models were not trained for this resolution, they have seen similar downscale resolution inside the network, but for a smaller input. The downscale inside the network will render correspondingly lower scale so each layer has not seen this particular scale at training. \cite{touvron2019} have shown that for the convolutional part of a CNN the receptive field is unaffected by the input size. We focus on the similarity between the models and assume that the workings of the models are still viable.

There are multiple datasets with real and synthetic image data. For our experiments, we chose BDD to represent a dataset of real images, along with GTAV to represent a dataset of synthetic images.

All models were initialized with the ImageNet pre-trained Darknet-53 backbone which populates layers 0 to 74. Layers 75 up to layer 106 was populated randomly according to Kai-Ming uniform distribution.

\begin{description}
    \item[U-Real] -- Further trained with all layers trainable (unfrozen) on our training set of BDD.
    \item[U-Synthetic] -- Further trained on the GTAV training set with all layers trainable (unfrozen).
    \item[F-Synthetic] -- Further trained on the GTAV training set with only detection head trainable i.e. layer 75-106 and thus leaving the backbone untrainable (frozen).
\end{description}

\subsection{Similarity Metric}
\label{sec:metric}
Comparing the similarity between two neural networks can be done in many ways. One approach is to look at the output for each individual layer and compare the outputs between networks. The problem can be described in the following way \citep{cka}:
\begin{quote}
Let $X_i\in \mathbb{R}^{p \times n}$ and $Y_i\in \mathbb{R}^{p \times n}$ be the output of layer $i$ in form of matrices from two networks with $p$ neurons each, fed with the same $n$ inputs. We want to introduce a metric function $s(X_i,Y_i)$ that can be used to compare the similarity between two output matrices, to give insight of the behaviour and similarities between the hidden layers inside the models.
\end{quote}

Several measures of similarity complying with this definition have been suggested. SVCCA \citep{svcca} and PWCCA \citep{pwcca} are two examples of measuring representational similarity. Both metrics are invariant to invertible linear transforms i.e.

\begin{equation}
    s(X, Y) = s(AX,BY)
    \label{eq:s}
\end{equation}

for any invertible matrices $A$ and $B$. This is argued to be an important property for comparing layer outputs. However, according to \cite{cka}, a metric with invariance to invertible linear transformations has the limitation of yielding the same similarity for all outputs with a greater width than the number of datapoints i.e. $p \geq n$.

The authors further argue that the scale of layer outputs also is important for representations. Therefore, similarity indices that preserve scale information, such as the Euclidean distance, can be helpful on giving insights of the activations. For a metric that is invariant to invertible transforms, the magnitude of the vectors in the activation space is irrelevant and therefore ignores this important information. 
Instead of requiring the similarity index to be invariant to invertible linear transform, a weaker invariance condition can be considered: invariance to orthogonal transformations. Invariance to orthogonal transformations means that $s(X,Y) = s(UX,VY)$ for any orthogonal matrices $U$ and $V$. A property is that invariance to orthogonal transformations also means invariance to permutations which is important since the convolutional layer outputs should have the same representations independent of channel-wise permutations. 

One such similarity index is linear CKA \citep{cka}. CKA is not only invariant to orthogonal transforms but also invariant to isotropic scaling i.e. $s(X,Y) = s(\alpha X, \beta Y)$ for any $\alpha, \beta \in \mathbb{R}^+$. 
For the matrices $X$ and $Y$, the CKA with a linear kernel is defined as:

\begin{equation}
    CKA(X,Y) = \frac{\|Y^T X\|^2_F}{\|X^T X\|_F \|Y^T Y\|_F},
    \label{eq:CKA}
\end{equation}

where $\|\cdot\|_F$ is the Frobenius norm and $n$ is the number of data points i.e. columns in $X$ and $Y$. With this index definition, \cite{cka} have shown that the CKA captures intuitive similarity ideas such as models trained in the same way with different initialization should be similar.

In our experiments, we used linear CKA.

\subsubsection{Convolutional Layers}
\label{sec:reshape}
While the CKA analysis requires matrices, the convolutional layers in the network output tensors. To solve this problem, we follow the line of \cite{cka} and treat the output tensors of shape $(n,h,w,c)$ as a collection of vectors of the shape $(n, h \cdot w \cdot c)$ where $n$ is the number of images fed through the network, $w$ and $h$ are the width and height of the image, and $c$ is the number of output channels (activations) i.e. the number of convolutional kernels for the specific layer.

\subsection{Representational Similarity}
Model U-Real gives us an indication of the performance we can obtain by only collecting a lot of real data.

Convolutional layers of the same layer index may have different roles in different networks trained on different data. Arguments can be made that the output of individual layers is not as important as the resulting output after a block of layers. However, here we focus on interpreting the single layer outputs.

The experiments used the CKA method described by \cite{cka} to analyze the similarity between layers of several models.

The layer-wise similarity analysis was done by feeding 200 random images from our BDD test set through the trained networks and performing CKA on the layer outputs to find which layers are similar and which are not.

Residual layers i.e. shortcut layers essentially just sum outputs from two layers without any weights, they are included in the CKA analysis for completeness of including all layers.

\subsection{Implementation}
The experiments were performed using the open-source implementation of YOLOv3 developed by \cite{ultralytics_yolov3}, using PyTorch 1.4 and the CKA implementation by \cite{cka}.

The performances presented as mean average precision (mAP) in the experiments are for all five common classes using mAP@0.5 i.e. mAP at 0.5 intersection over union (IoU).


\section{Results}
\label{sec:results}
In order to see that the trainings were successful, the resulting mAP of the trained models evaluated on our test set of the synthetic GTAV dataset and our test set of the real BDD dataset using image size $416 \times 416$ are presented in Table \ref{tab:synth_performance_test_paper}.

\begin{table*}[t]
 \caption{Performance of models trained on GTAV synthetic data: U-Synthetic and F-Synthetic as well as model U-Real trained on our BDD training set. All evaluated for mAP on our GTAV test set and our BDD test set.}
    \begin{center}
    \begin{tabular}{ |p{2.0cm}||p{1.5cm}|p{1.5cm}|p{1.5cm}|p{1.5cm}|}
    \hline
\textbf{Model} & \multicolumn{2}{c|}{\textbf{mAP on BDD}} & \multicolumn{2}{c|}{\textbf{mAP on GTAV}} \\
 & seed 0 & seed 1 & seed 0 & seed 1 \\
\hline
  U-Real        & 0.428 & 0.430 & 0.440 & 0.439 \\
  U-Synthetic   & 0.122 & 0.124 & 0.886 & 0.893 \\
  F-Synthetic   & 0.125 & 0.121 & 0.892 & 0.884 \\
\hline
\end{tabular}
\end{center}
\label{tab:synth_performance_test_paper}
\end{table*}

The models yielded best mAP on the type of data they were trained for, where U-Real got about 0.43 mAP on BDD while model U-Synthetic and F-Synthetic both only got about 0.12 mAP. Tested on GTAV model U-Synthetic and F-Synthetic both got about 0.89 mAP while U-Real got about 0.44 mAP.

It can be seen that model U-Synthetic and F-Synthetic had comparable mAP on both BDD and GTAV respectively, considering variations of trainings with different random seeds, see Table \ref{tab:synth_performance_test_paper}.

\subsection{Layer-wise Analysis}

\begin{figure*}[ht]
    \centering
    \includegraphics[width=\linewidth]{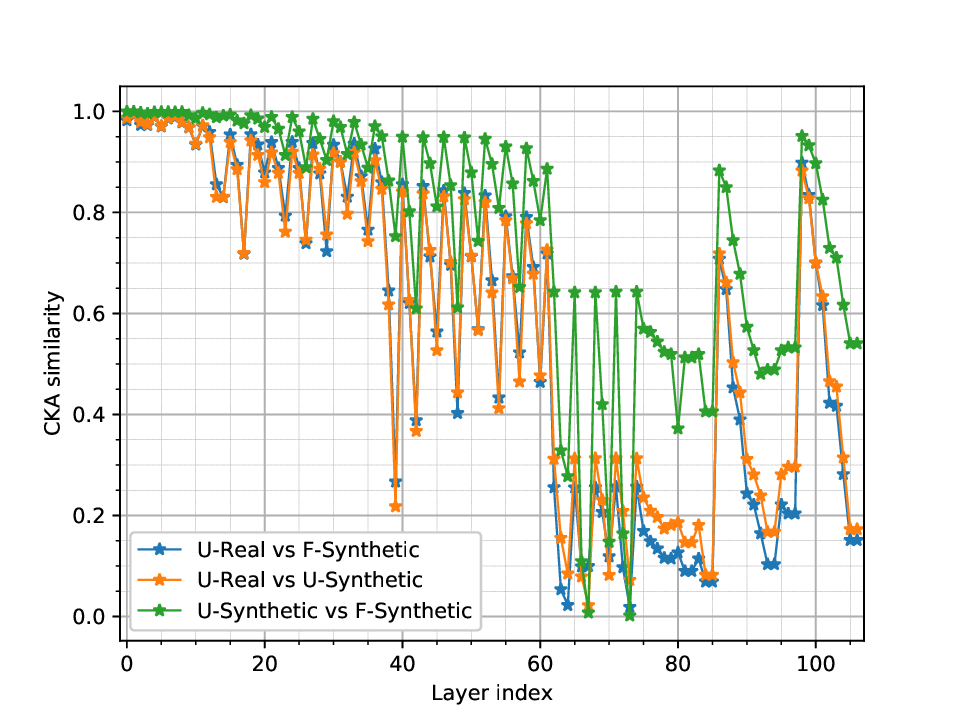}
    \caption{CKA similarity for all layers in YOLOv3 when images from our BDD test set were passed through the networks that were initialised with seed 0 and trained.}
    \label{fig:cka_layer_0_to_106_all_layers}
\end{figure*}

\begin{figure*}[ht]
    \centering
    \includegraphics[width=\linewidth]{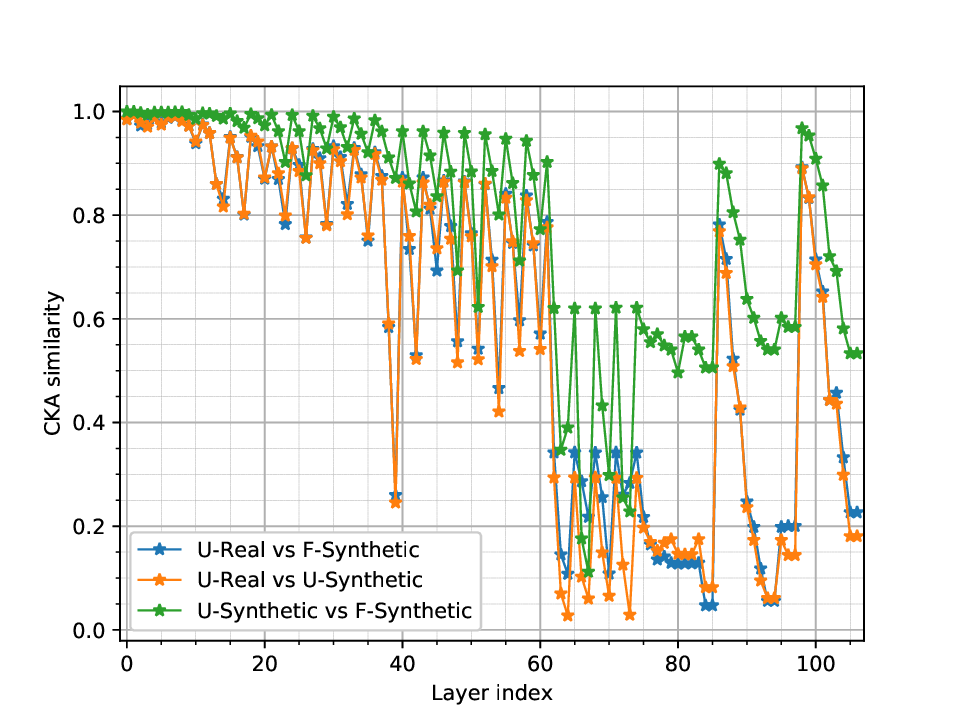}
    \caption{CKA similarity for all layers in YOLOv3 when images from our BDD test set were passed through the networks that were initialised with seed 1 and trained.}
    \label{fig:cka_layer_0_to_106_all_layers_seed1}
\end{figure*}

The objective was to observe differences in models trained on real and synthetic data. Model U-Real was trained on real data only (ImageNet + BDD), while the head parts of U-Synthetic and F-Synthetic were trained on synthetic data only. Figures \ref{fig:cka_layer_0_to_106_all_layers} and \ref{fig:cka_layer_0_to_106_all_layers_seed1} show the results of the CKA similarity analysis using 200 images from our BDD test set that were fed through the models.

Summary statistics of all layer outputs (activations), averaged over all layers, are presented in Tables \ref{tab:summary_stats} and \ref{tab:summary_stats32}. A small difference in distribution can be observed between model U-Real trained on real data and the models trained on synthetic data: Models U-Synthetic and F-Synthetic. The difference was mostly in the mean and standard deviation. Comparably, models U-Synthetic and F-Synthetic have quite similar layer output value distribution. This can be observed both for image size $416 \times 416$ and $32 \times 32$, making it consistent between the mAP analysis and CKA analysis.

\begin{table*}[t]
 \caption{Summary statistics of all layer outputs when feeding the network with 200 images of size $416 \times 416$ from our BDD test set. Values were averaged over all layers.}
    \begin{center}
    \begin{tabular}{|p{2.0cm}||p{0.7cm}|p{1.3cm}|p{1.3cm}|p{1.0cm}|p{1.0cm}|p{1.0cm}|}
    \hline
\textbf{Model} & seed & \textbf{mean} & \textbf{median} & \textbf{std} & \textbf{min} & \textbf{max}\\
\hline
  U-Real        & 0 & -0.0165 & -0.116 & 0.961 & -21.8 & 54.4 \\
  U-Real        & 1 & -0.0145 & -0.113 & 0.959 & -21.0 & 52.1 \\
  U-Synthetic   & 0 & -0.0272 & -0.113 & 0.986 & -22.0 & 52.4 \\
  U-Synthetic   & 1 & -0.0269 & -0.110 & 0.995 & -22.3 & 50.3 \\
  F-Synthetic   & 0 & -0.0263 & -0.111 & 1.00  & -22.2 & 51.0 \\
  F-Synthetic   & 1 & -0.0235 & -0.110 & 0.983 & -23.2 & 48.0 \\
\hline
\end{tabular}
\end{center}
\label{tab:summary_stats}
\end{table*}

\begin{table*}[t]
 \caption{Summary statistics of all layer outputs when feeding the network with 200 images of size $32 \times 32$ from our BDD test set. Values were averaged over all layers.}
    \begin{center}
    \begin{tabular}{|p{2.0cm}||p{0.7cm}|p{1.3cm}|p{1.3cm}|p{1.0cm}|p{1.0cm}|p{1.0cm}|}
    \hline
\textbf{Model} & seed & \textbf{mean} & \textbf{median} & \textbf{std} & \textbf{min} & \textbf{max}\\
\hline
  U-Real        & 0 & 0.0613 & -0.105 & 0.812 & -16.8 & 36.8 \\
  U-Real        & 1 & 0.0634 & -0.107 & 0.817 & -17.0 & 33.8 \\
  U-Synthetic   & 0 & 0.0909 & -0.103 & 0.743 & -15.3 & 40.1 \\
  U-Synthetic   & 1 & 0.0989 & -0.104 & 0.720 & -15.8 & 37.3 \\
  F-Synthetic   & 0 & 0.0996 & -0.105 & 0.726 & -15.0 & 38.2 \\
  F-Synthetic   & 1 & 0.0925 & -0.102 & 0.731 & -15.2 & 39.5 \\
\hline
\end{tabular}
\end{center}
\label{tab:summary_stats32}
\end{table*}

\subsubsection{CKA Similarity Analysis}
The CKA similarity can be seen in Figures \ref{fig:cka_layer_0_to_106_all_layers} and \ref{fig:cka_layer_0_to_106_all_layers_seed1} for each seed respectively. There was high similarity between real and synthetic models for both U-Synthetic and F-Synthetic in the first 13 layers of the network where all have similarity above 0.9 and most of them above 0.95. The similarity was above 0.7 for the first layers until layer 37. After layer 37 there was more variation in the similarity between the models. The similarity was quite high in most of the backbone until layer 61.

Comparing model U-Real with the synthetic models, the similarity from layer 62 to 85 was under 0.35. The similarity was relatively low for the three detection layers in the head (layers 82, 94, 106), including their preceding layer, where all have similarity between 0.05 and 0.2.

Frequent ups and downs in the similarity can be seen, where all the local peaks in the backbone corresponds to the residual layers, for both seeds.

In the head part, there were two peaks, at route layers 86 (concatenating the previous layer output with the output from layer 61) and 98 (concatenating the previous layer output with the output from layer 36), both routing from layers in the backbone. Since the similarity was high in the backbone overall, it is reasonable that there were similarity peaks where those two backbone layers are routed in the head part.

In the head part, each detection layer and the one immediately preceding convolutional layer had the same similarity values.

The average of CKA similarity was higher in the backbone than in the head part for both comparisons of U-Real vs the synthetic models, see Table \ref{tab:mean_cka}. Likewise, the similarity between model U-Synthetic and model F-Synthetic was overall higher than when compared to model U-Real. The head part had lower similarity than the backbone and lower than the mean of all layers, for all comparisons.

Note the part between layer 62 and 85 that all had lower similarity values than the rest of the network in all comparisons, for both seeds. This region corresponds to the lowest scale in the network.

The input images of size $32 \times 32$ have the size $1 \times 1$ in this region, which was lower than the convolutional kernel of $3 \times 3$ used in most layers in the entire network. However, for larger image sizes, it was not possible to perform these CKA calculations for the entire network since it would demand a vast amount of working memory. However, they could be performed for large parts of the network and larger image input sizes such as $64 \times 64$ and $128 \times 128$ showed similar patterns in that region, see Section \ref{layer_vs_layer}.

No difference was found for the U-Synthetic unfrozen model or the F-Synthetic frozen model in terms of overall average similarity with the unfrozen model U-Real, considering trainings with different random seeds, see Table \ref{tab:mean_cka}. Thus, there was no overall impact of frozen or unfrozen in this regard.

\begin{table*}[t]
 \caption{Mean CKA similarity for the models, for all layers, backbone and head.}
    \begin{center}
    \begin{tabular}{|p{3.3cm}||p{1.0cm}|p{1.0cm}|p{1.0cm}|p{1.0cm}|p{1.0cm}|p{1.0cm}|}
    \hline
\textbf{Model} & \multicolumn{2}{c|}{\textbf{all}} & \multicolumn{2}{c|}{\textbf{backbone}} & \multicolumn{2}{c|}{\textbf{head}} \\
                  & seed 0 & seed 1 & seed 0 & seed 1 & seed 0 & seed 1 \\
\hline
  U-Real vs U-Synth. & 0.5865 & 0.5895 & 0.6925 & 0.7112 & 0.3379 & 0.3042 \\
  U-Real vs F-Synth. & 0.5734 & 0.6054 & 0.6931 & 0.7318 & 0.2930 & 0.3092 \\
  U-Synth. vs F-Synth. & 0.7597 & 0.7845 & 0.8264 & 0.8461 & 0.6115 & 0.6508 \\
\hline
\end{tabular}
\end{center}
\label{tab:mean_cka}
\end{table*}

\subsubsection{Different seeds}
\begin{figure*}[ht]
    \centering
    \includegraphics[width=\linewidth]{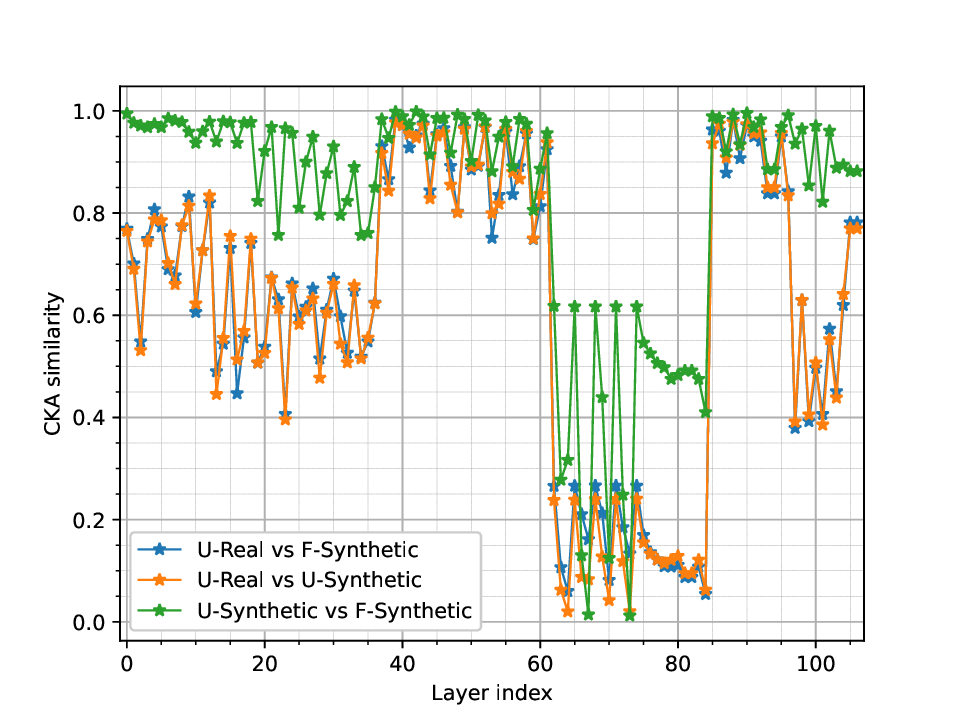}
    \caption{CKA similarity for all layers in YOLOv3 when images from our BDD test set were passed through the trained models with different initialization; comparing seed 0 with seed 1.}
    \label{fig:cka_layer_0_to_106_all_layers_3models_seed0_seed1}
\end{figure*}

Comparing the similarity of using different seeds for initialization, the CKA similarity for the models trained with seed 0 compared to trained with seed 1 can be seen in Figure \ref{fig:cka_layer_0_to_106_all_layers_3models_seed0_seed1}. A similar structure can be seen as in the above results in Figures \ref{fig:cka_layer_0_to_106_all_layers} and \ref{fig:cka_layer_0_to_106_all_layers_seed1} where the same seed was used in each comparison. However, some differences to the plots with the same seed can be seen. For example, not all the residual layers had peaks and when comparing the U-Real model with the synthetic models it can be seen that the early layers did not have similarity close to 1 while the similarity was higher in the end of the network.

\subsection{Layer vs Layer Analysis}
\label{layer_vs_layer}
\begin{figure*}[ht]
    \centering
    \includegraphics[width=\linewidth]{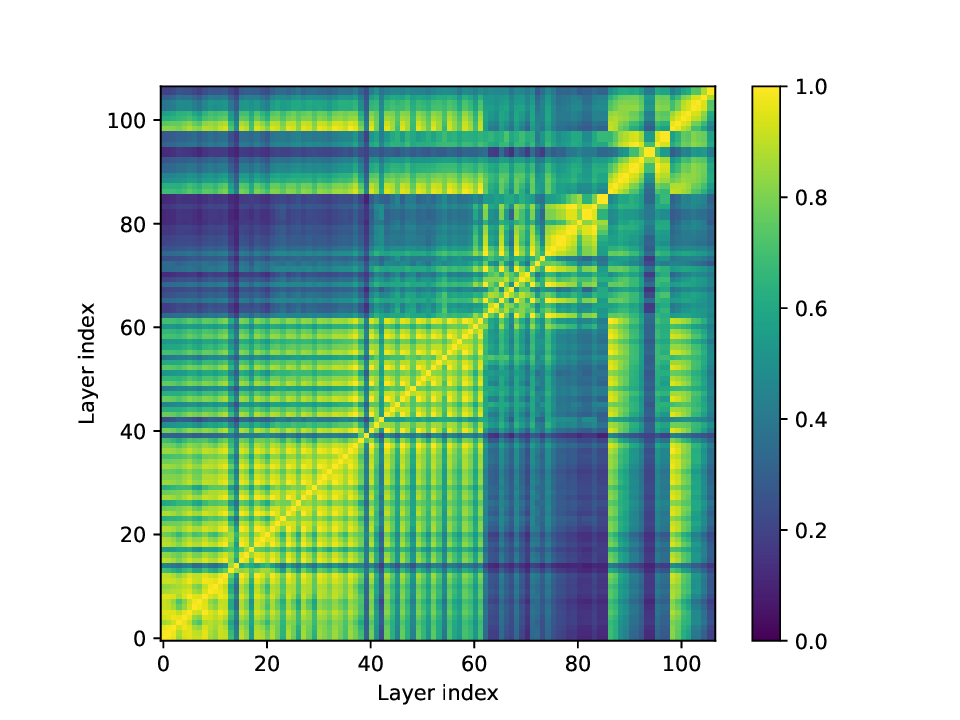}
    \caption{CKA similarity between layers of model U-Real that was initialised with seed 0 and trained, for all layers in YOLOv3.}
    \label{fig:cka_layer_vs_layer_b_vs_b}
\end{figure*}

Looking at CKA similarity between layers shows how each layer compares to all other layers in the network, see Figures \ref{fig:cka_layer_vs_layer_b_vs_b}, \ref{fig:cka_layer_vs_layer_b_vs_c}, \ref{fig:cka_layer_vs_layer_b_vs_d}, \ref{fig:cka_layer_vs_layer_c_vs_d}, \ref{fig:cka_layer_vs_layer_b_vs_c_imgsize64} and \ref{fig:cka_layer_vs_layer_c_vs_b_imgsize128}.

A row in these plots consists of the CKA similarity values between one layer in the model on the $y$-axis and all layers in the model on the $x$-axis.

A grid-like structure was visible for different parts of the YOLOv3 architecture and the different trained models individual characteristics. Layers 0 to 12 were mostly similar to each other within the same model, as well as with other models, in all comparisons. Regions can be seen approximately for layers 0 to 12, 14 to 37, 42 to 61, 62 to 74, 75 to 84, 85 to 96, and 97 to 106. This represents the architectural structure of YOLOv3 with approximate scale regions.

As could be seen in the CKA plot in Figures \ref{fig:cka_layer_0_to_106_all_layers} and \ref{fig:cka_layer_0_to_106_all_layers_seed1}, the impact of the routing layers (86 and 98) in the head part can be seen near the diagonal. The part with the maximum downscale between layers 62 and 85 can be seen here as well, this part had lower similarity with most other layers in the network.

Figure \ref{fig:cka_layer_vs_layer_b_vs_b} shows the similarity of all the layers against each other in model U-Real; self-similarity symmetric around the diagonal.

\subsubsection{Real vs Synthetic Data}
\begin{figure*}[ht]
    \centering
    \includegraphics[width=\linewidth]{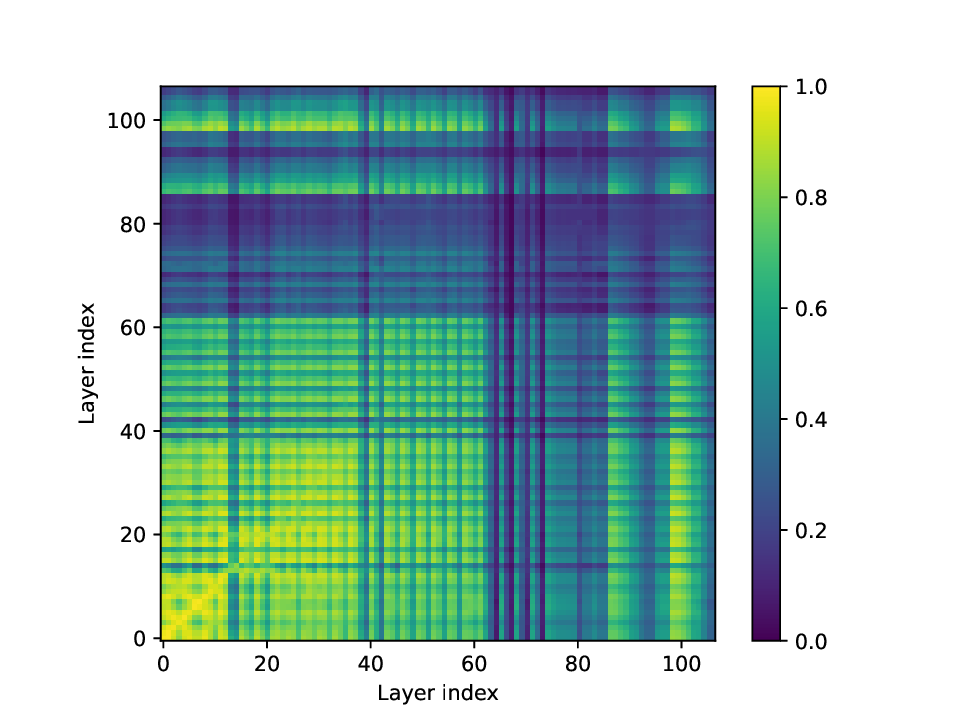}
    \caption{CKA similarity between layers of model U-Real ($y$-axis) vs model U-Synthetic ($x$-axis) that were initialised with seed 0 and trained, for all layers in YOLOv3.}
    \label{fig:cka_layer_vs_layer_b_vs_c}
\end{figure*}

\begin{figure*}[ht]
    \centering
    \includegraphics[width=\linewidth]{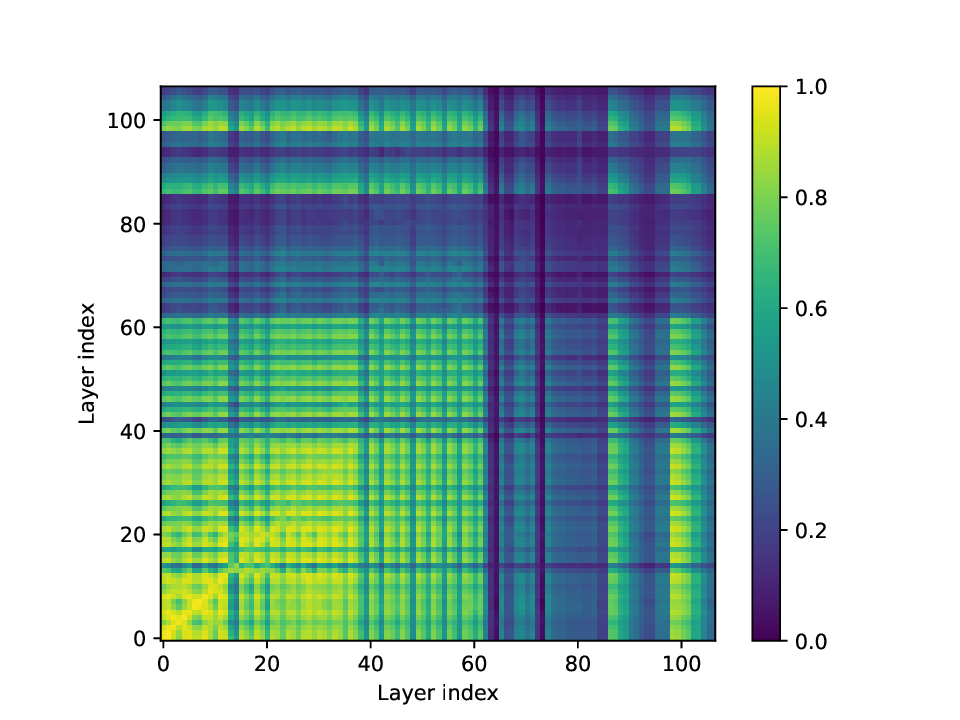}
    \caption{CKA similarity between layers of model U-Real ($y$-axis) vs model F-Synthetic ($x$-axis) that were initialised with seed 0 and trained, for all layers in YOLOv3.}
    \label{fig:cka_layer_vs_layer_b_vs_d}
\end{figure*}

The diagonals of the plots of model U-Real vs U-Synthetic, U-Real vs F-Synthetic, and U-Synthetic vs F-Synthetic, seen in Figures \ref{fig:cka_layer_vs_layer_b_vs_c}, \ref{fig:cka_layer_vs_layer_b_vs_d}, and \ref{fig:cka_layer_vs_layer_c_vs_d}, are the same as the curves seen in Figure \ref{fig:cka_layer_0_to_106_all_layers}. The values off the diagonal thus show the similarity of layers with differing layer numbers. 

The last layers in the backbone differs between comparisons of U-Real vs U-Synthetic, and U-Real vs F-Synthetic, seen in Figures \ref{fig:cka_layer_vs_layer_b_vs_c}, \ref{fig:cka_layer_vs_layer_b_vs_d}.

Most of the differences for real and synthetic were between layer 62 and 85, where the features were downscaled to the lowest scale. There U-Real seems more similar to layers in U-Synthetic and F-Synthetic, than U-Synthetic and F-Synthetic were similar to layers in U-Real.

\subsubsection{Frozen vs Unfrozen Backbone}
\begin{figure*}[ht]
    \centering
    \includegraphics[width=\linewidth]{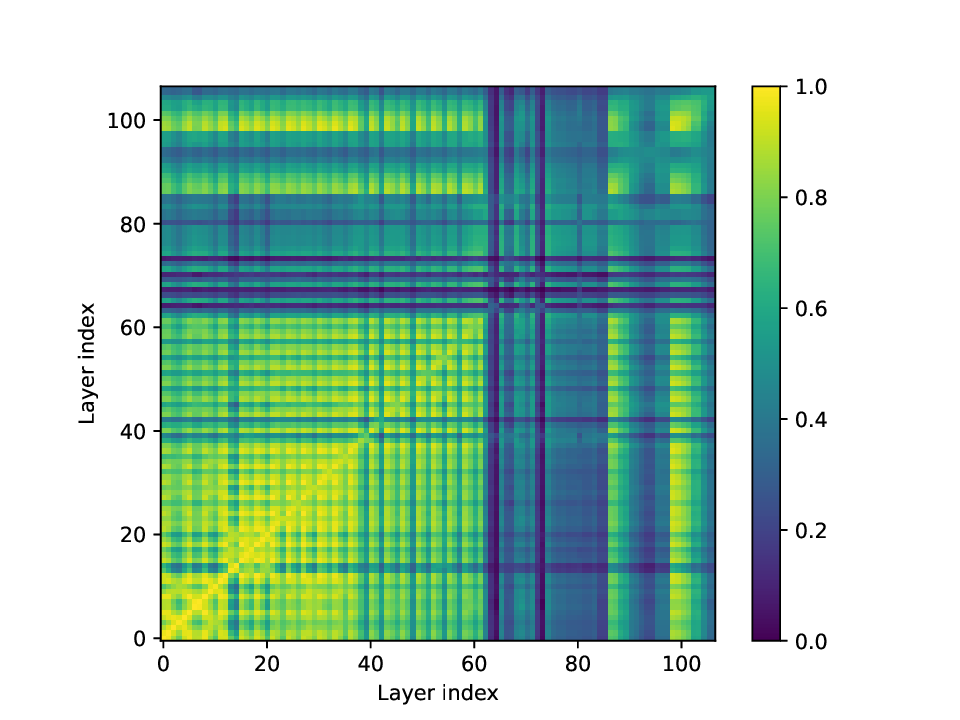}
    \caption{CKA similarity between layers of model U-Synthetic ($y$-axis) vs model F-Synthetic ($x$-axis) that were initialised with seed 0 and trained, for all layers in YOLOv3.}
    \label{fig:cka_layer_vs_layer_c_vs_d}
\end{figure*}

Model U-Synthetic and F-Synthetic similarity to model U-Real for each layer can be seen in Figures \ref{fig:cka_layer_vs_layer_b_vs_c} and \ref{fig:cka_layer_vs_layer_b_vs_d}. The similarity between all layers in model U-Synthetic (unfrozen) and F-Synthetic (frozen) can be seen in Figure \ref{fig:cka_layer_vs_layer_c_vs_d}. In comparison with the similarity plots of U-Real vs U-Synthetic, and U-Real vs F-Synthetic, the similarity between U-Synthetic and F-Synthetic was overall higher. There was high similarity in most of the backbone, specially the first part, even though model U-Synthetic had trainable backbone and model F-Synthetic had frozen backbone. However, a few layers in the backbone differs, for example the last layers in the backbone. The largest differences were thus in the head part, except for higher similarity around the route layers 86 and 98.

\subsubsection{Other Input Resolutions}
\begin{figure*}[ht]
    \centering
    \includegraphics[width=\linewidth]{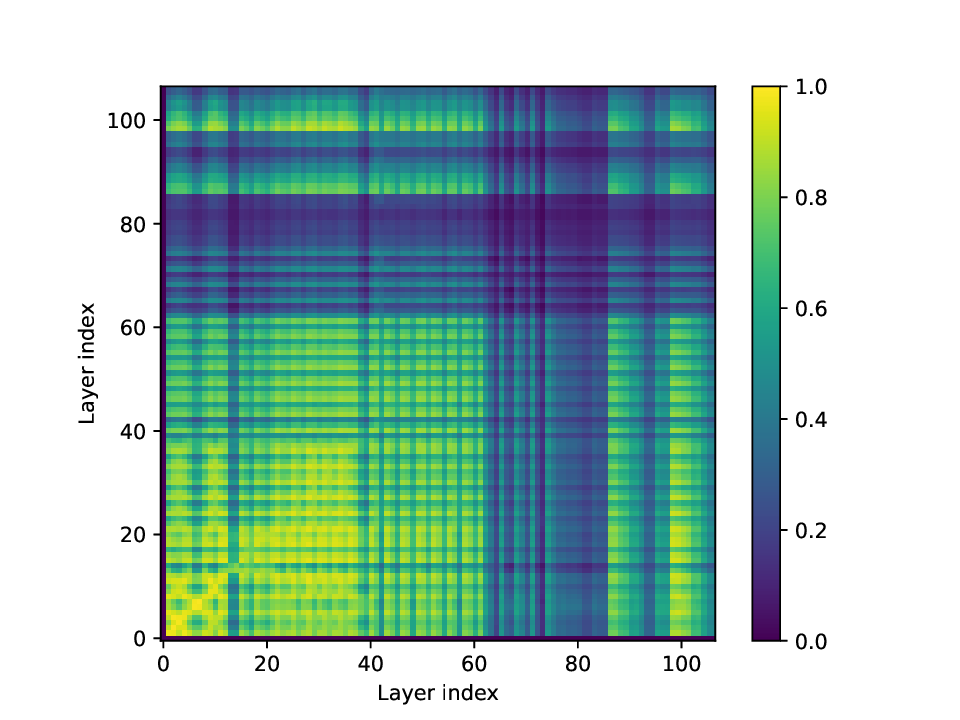}
    \caption{CKA similarity between layers of model U-Real ($y$-axis) vs model U-Synthetic ($x$-axis) that were initialised with seed 0 and trained, input size of $64 \times 64$. Results for layers 2 to 106 in YOLOv3.}
    \label{fig:cka_layer_vs_layer_b_vs_c_imgsize64}
\end{figure*}

\begin{figure*}[ht]
    \centering
    \includegraphics[width=\linewidth]{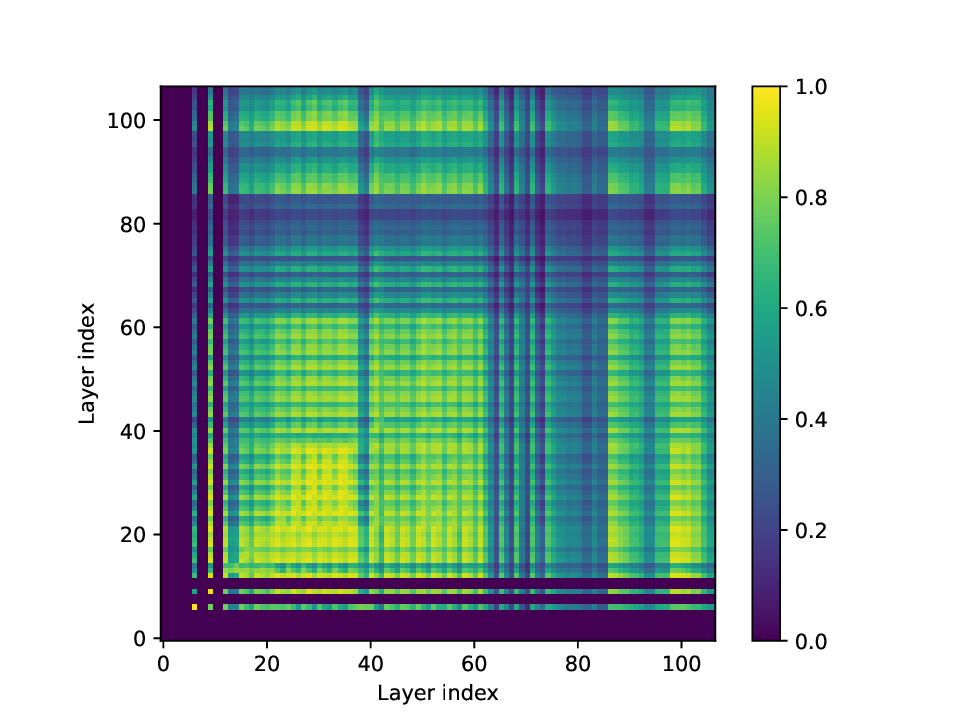}
    \caption{CKA similarity between layers of model U-Real ($y$-axis) vs model U-Synthetic ($x$-axis) that were initialised with seed 0 and trained, input size of $128 \times 128$. Results for layers 6, 9, 12 to 106 in YOLOv3.}
    \label{fig:cka_layer_vs_layer_c_vs_b_imgsize128}
\end{figure*}

Larger image input sizes such as $64 \times 64$ and $128 \times 128$ showed similar structure as the above mentioned results for $32 \times 32$, see Figures \ref{fig:cka_layer_vs_layer_b_vs_c_imgsize64} and \ref{fig:cka_layer_vs_layer_c_vs_b_imgsize128}.

\section{Discussion}
\label{sec:discussion}

The results overall showed small differences between model U-Real trained on real data and the models trained on synthetic data: Models U-Synthetic and F-Synthetic. The first part of the models showed high CKA similarity in all comparisons, while the head part showed more differences. All models had the same backbone pre-trained on real image data from ImageNet and model F-Synthetic did not have further training of the backbone. The high similarity between all models in the backbone means that the pre-trained backbone is rather dominant in all models, even after further training of the backbone in model U-Real and model U-Synthetic.

The similarity between model U-Synthetic and model F-Synthetic was higher than when these models were compared to model U-Real. It seems like models trained on the same dataset develop similar representations. However, it could be further explored how much of this is due to real vs synthetic data and different datasets in general.

\subsection{First Layers}
The first 13 layers in the backbone had very high similarity between all models, with similarity well above 0.9, when using the same seed. Thus, the first layers in the network were not affected much by the dataset type and are likely mostly from the pre-trained backbone. These layers are likely targeting generic image features, such as edges in the first layer \cite{goodfellow2016}. All models had the same backbone pre-trained on real image data from ImageNet, but that does not explain why the first 13 layers would be more similar than the rest of the backbone. The high similarity shows that the early layers of the different models develop similar representations, irrespective of if the dataset is real or synthetic.

\subsection{Residual Layers}
There were also frequent ups and downs in the similarity. All of the local peaks in the backbone were for the residual layers, when using the same seed. Residual layers sum the output of the current layer and a previous layer, so information is partly propagated through the network. It could happen that residual blocks are trained to have very small weights, which would have the result that the information from the previous layer (the shortcut) propagates quite unaffected \cite{ResNet,identitymappings}. Since the first 13 layers had very high CKA similarity, this could be why the residual layers also have high CKA similarity. The convolutional layers between the residual layers had a bit lower similarity and were thus more unique.

\subsection{Different Seeds}
Comparing the similarity of using different seeds for initialization, it showed similarities with using the same seed, but not all the residual layers had peaks and when comparing the U-Real model with the synthetic models it can be seen that the early layers did not have similarity close to 1, while the similarity was higher in the end of the network. This indicates that the similarity can be calculated for different training seeds, as showed in \cite{cka}, but that there can be some variation still, as also showed in \cite{nguyen2021cka}.

\subsection{Frozen vs Unfrozen Backbone}
The F-Synthetic model with frozen backbone and the U-Synthetic model with unfrozen backbone, both further trained on the synthetic GTAV dataset, both had comparable mAP on BDD and GTAV respectively. No particular difference could be seen in the CKA analysis between frozen and unfrozen backbone. In \cite{hinterstoisser2018buzz}, freezing the backbone during training on synthetic images yielded better performance on a real dataset compared to using an unfrozen backbone. However, \cite{tremblay2018nvidia} showed promising results for unfrozen backbone. The diversity of the domain randomized dataset that they used could be the explanation to why they find differing results. To summarize, it seems that there is not a consensus whether freezing the backbone or not is preferred in all cases.

Comparing model U-Synthetic with unfrozen backbone and model F-Synthetic with frozen backbone, there were high similarity in most of the backbone between the two, specially the first part. However, a few layers in the backbone differ, for example the last layers in the backbone. Both models were derived from the same pre-trained backbone and perhaps the training of model U-Synthetic with trainable backbone did not result in large updates in the backbones.

\subsection{Middle Layers}
In all comparisons, CKA similarity was lower than the rest of the network in the part between layer 62 and 85. The features were downscaled successively in the network and between layer 62 and 85 they have the smallest size. For input images of size $32 \times 32$ the size of these layers become $1 \times 1$, for input image size $64 \times 64$ the size is $2 \times 2$, and for input image size $128 \times 128$ the layer size is $4 \times 4$. The CKA similarity was lower in this region for all input image sizes tested so it is not sure that the particular small size of $1 \times 1$ in those layers would explain this, but the relative downscaling could perhaps have a general effect. This low-scale region of the network was presumably trained to detect prominent or spatially large feature structures.

\subsection{Head Part}
In the head part to the right, the three detection layers 82, 94 and 106, we see that the similarity is very low, below 0.2 when using the same seed. This implies that these layers are quite unique between the three models.

The largest differences between model U-Synthetic and F-Synthetic were in the head part. Since model U-Synthetic had trainable backbone while model F-Synthetic had frozen backbone it would be expected that their backbones differ. Both networks were trained on the same detection task on the same dataset, so the head parts could likely become similar due to that. However, the head parts integrate information from multiple layers in the backbone that all have differences. Also, the receptive field increases with layer number and thus is quite large in the head part. These factors may explain why the largest differences were in the head part.

\subsection{Comparison With Previous Work}
\cite{cka} applied image classification on two different datasets with real images of resolution $32 \times 32$ using a 9 layer CNN network. The CKA similarity between the trained models was close to 1 for all comparisons for layers 1-4 irrespective of dataset, then dropped somewhat for later layers, especially after about layer 6. Similarity between the trained and untrained models was about 0.8 for the first layer and then dropped in a slope towards near zero for the last layer. This implies that a CKA similarity value of 0.8 could mean that the first layer of the trained model, which usually targets generic image features, was somewhat similar to random noise. In another experiment with two untrained models with different initializations, the CKA similarity of the first layer was near 1 and for the first couple of layers were about 0.8 approximately. Our results are consistent with these results in that the early layers of the models showed high similarity, in our case mostly above 0.9, also somewhat for different initializations.

Higher CKA similarity values mean high similarity and vice versa, but in between high and low it is not entirely clear how different CKA similarity values should be interpreted.

\cite{nguyen2021cka} showed further analyses of CKA on different ResNet architectures for image classification. Since the backbone of YOLOv3 has similarities with ResNet, our analysis showed similar results in general. They found a block structure of deep models having large blocks with high similarity - indicating overparameterisation in relation to the training data. However, our results did not show large blocks of very high similarity, only very local similarities between layers like $3 \times 3$ and some larger area but with varying similarity. Our plots did not show these large high similarity blocks, probably since we did have large amounts of training data.

Comparing CKA similarity values for layers vs layers showed the differences between the models individual characteristics, similar to the results in \cite{nguyen2021cka}.

\subsection{Limitations and Future Work}
Here we trained on image size $416 \times 416$ while analysing CKA on image size $32 \times 32$ which is a scale that the models were not trained for, which is a limitation, but we focus on the similarity between the models. Furthermore, in this work, one network architecture was analysed and trainings using one real image dataset with one synthetic image dataset were compared.

In future work, the analysis would benefit of looking at multiple real and synthetic datasets and compare them as groups. To further study the differences between training on real and synthetic data it would be of interest to investigate the impact of photorealism. Furthermore, different network architectures and tasks such as segmentation could also be analysed.

\section{Conclusions}
\label{sec:conclusions}

In our paper, we dissected models trained on real and synthetic images. We started from a backbone pre-trained on ImageNet real image data. Then:
\begin{itemize}
    \item One model, U-Real, was further trained on real image data (BDD).
    \item Two other models were further trained on synthetic data (GTAV):
    \begin{itemize}
        \item Model U-Synthetic with all layers trainable (unfrozen), and
        \item Model F-Synthetic with a frozen backbone.
    \end{itemize}
\end{itemize}  

Our main contributions are twofold. We show what parts of the network were most and least similar comparing a detector trained on real image data to when it was trained on synthetic data. We also show that freezing the backbone or not does not have to be important when further training a detector on synthetic data.

The trained models yielded best mAP on the type of data they were trained for.

Summary statistics of all layer outputs showed a small difference in distribution between model U-Real trained on real data and the models trained on synthetic data; models U-Synthetic and F-Synthetic. Comparably, models U-Synthetic and F-Synthetic have quite similar layer output value distribution.

The CKA similarity was calculated for comparing the model trained on real data, model U-Real, with models trained on synthetic data, models U-Synthetic and F-Synthetic.
The average CKA similarity was higher in the backbone than in the head part when comparing the model trained on real data with the two models trained on synthetic data. Specially the first 13 layers in the backbone had high similarity between all models, thus the first layers in the network were not affected much by the dataset type.

The similarity was quite high in most of the backbone until layer 61. From layer 62 to 85, the feature size was the lowest and the similarity was relatively low.

The head part had lower similarity than the backbone, which was also lower than the mean of all layers. The similarity was relatively low for the three detection layers in the head.

Comparing CKA similarity values for layers vs layers showed a grid-like structure resembling the different parts of the YOLOv3 architecture and also the differences between the models individual characteristics could be seen.

No major difference could be seen in either mAP or the CKA analysis between frozen and unfrozen backbone. However, the largest difference between model U-Synthetic, with unfrozen backbone, and model F-Synthetic, with frozen backbone, according to CKA was in the head part. Hence they were more similar to each other in the backbone part than in the head part, even though their backbones had different training settings.

With this similarity analysis, we want to give insights on how training synthetic data affects each layer and to give a better understanding of the inner workings of complex neural networks. A better understanding is a step towards using synthetic data in an effective way and towards explainable and trustworthy models.

\backmatter


\bmhead{Acknowledgments}
Thanks to our colleagues for valuable discussions.

The Version of Record of this article is published in Springer Nature Computer Science 2023, and is available online at
\href{https://doi.org/10.1007/s42979-023-01704-5}{doi}

\section*{Declarations}
\subsection*{Conflict of interest}
The authors declare that they have no conflict of interest.

\section*{Data availability}
The dataset splits and links to the datasets are available on \href{https://github.com/ljungqvistmartin/datasplits}{github}

The CKA code by Kornblith et al. is available on \href{https://github.com/google-research/google-research/tree/master/representation_similarity}{github}

\bibliography{references}


\end{document}